\documentclass{article}

\PassOptionsToPackage{numbers, compress}{natbib}

\usepackage[preprint]{neurips_2024}

\usepackage[utf8]{inputenc} 
\usepackage[T1]{fontenc}    
\usepackage{url}            
\usepackage{booktabs}       
\usepackage{amsfonts}       
\usepackage{nicefrac}       
\usepackage{microtype}      

\usepackage{floatrow}
\usepackage{wrapfig}
\usepackage{moresize}
\usepackage[font={small}]{caption}
\usepackage{numprint}
\usepackage{scalerel}

\usepackage{animate}
\usepackage{booktabs}
\usepackage{multirow}
\usepackage{cancel}
\usepackage{soul}
\usepackage{subcaption}
\usepackage{tabularx}
\usepackage{enumitem}
\usepackage[dvipsnames]{xcolor}
\usepackage{pifont}
\newcommand{\cmark}{\ding{51}}%
\newcommand{\xmark}{\ding{55}}%
\newcommand{\supplementary}{\textcolor{RoyalBlue}{Appendix}}

\newcommand{\eg}{\textit{e}.\textit{g}.}

\usepackage[strict]{changepage}

\usepackage{framed}

\definecolor{formalshade}{rgb}{0.95,0.95,0.95}

\usepackage[pagebackref=true,breaklinks=true,letterpaper=true,colorlinks,bookmarks=false]{hyperref}
\title{\vspace{-0.5cm}\raisebox{-0.7ex}{\includegraphics[scale=0.04]{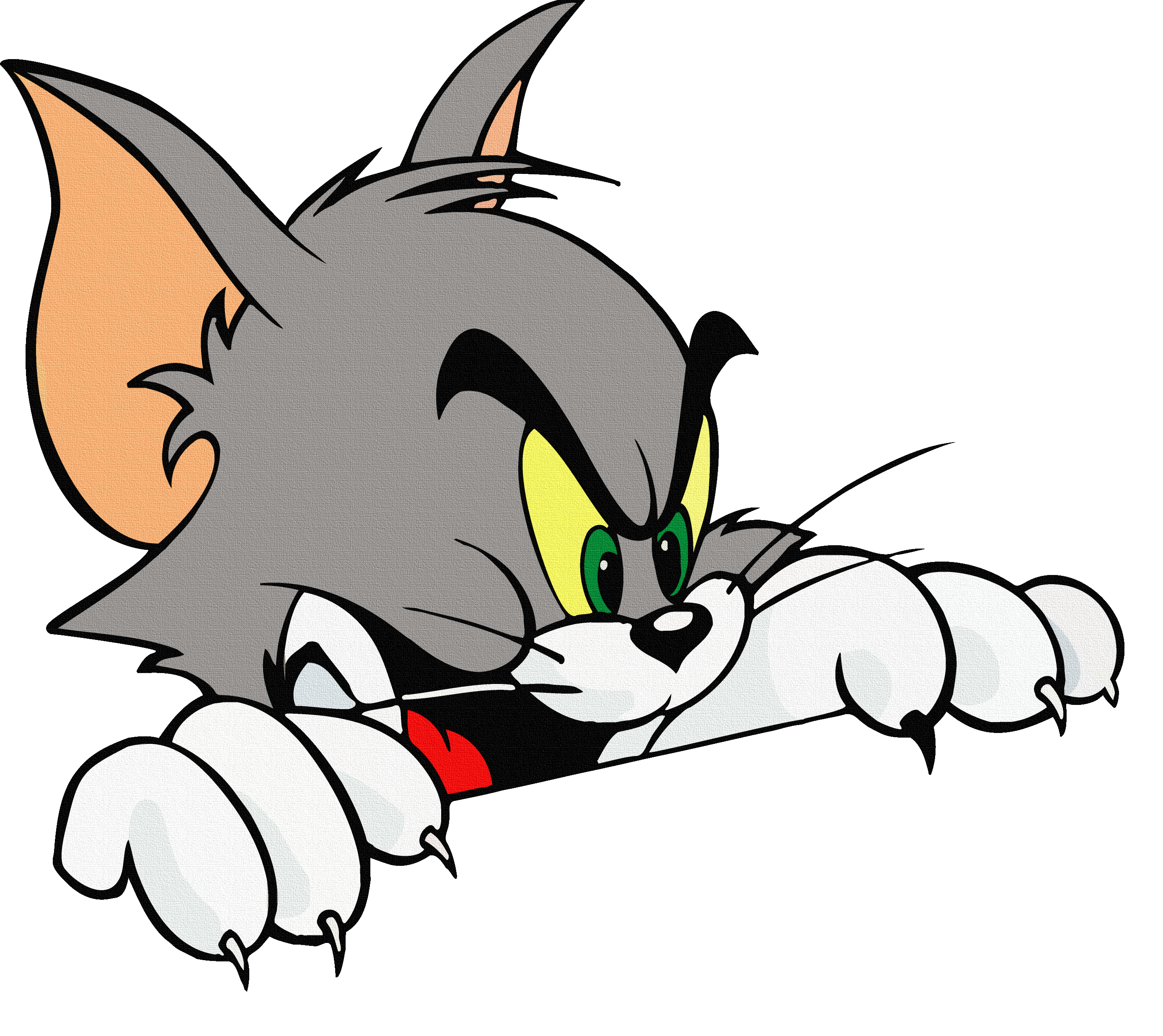}} \textit{CausalChaos!} Dataset for Comprehensive Causal Action Question Answering Over Longer Causal Chains Grounded in Dynamic Visual Scenes\raisebox{-0.2ex}{\includegraphics[scale=0.02]{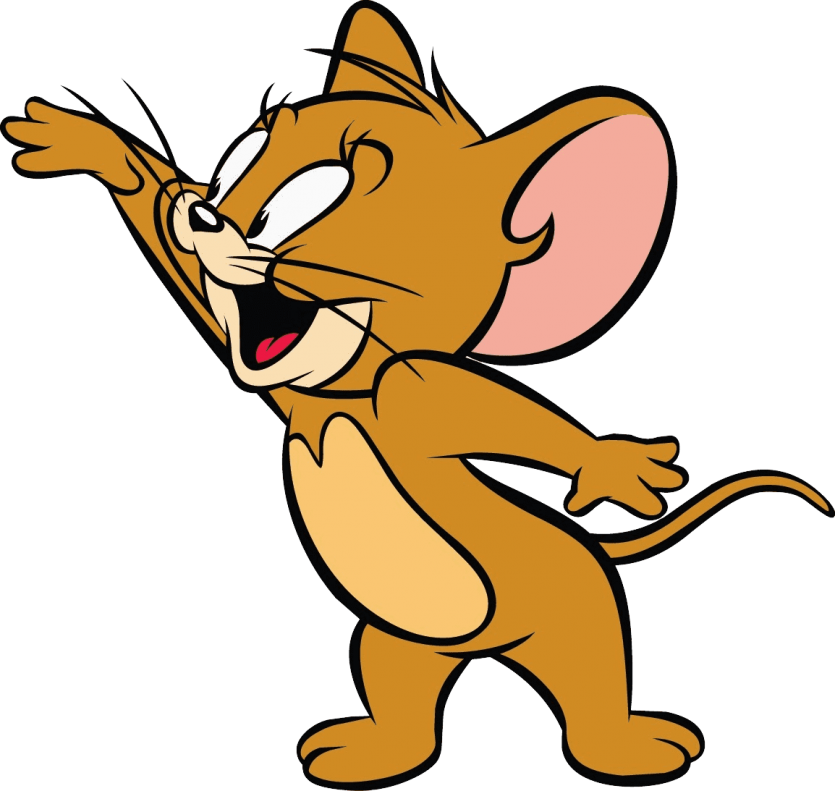}}}

\author{%
  Paritosh Parmar$^{1}$, Eric Peh$^{1}$, Ruirui Chen$^{1}$, Ting En Lam$^{1,2}$, \\ \textbf{Yuhan Chen$^{1,3}$, Elston Tan$^{1,4}$, Basura Fernando$^{1}$}\\
\small{$^{1}$Institute of High-Performance Computing, Agency for Science, Technology and Research, Singapore}\\
\small{$^{2}$Nanyang Technological University,   $^{3}$National University of Singapore, $^{4}$Singapore Polytechnic}\\
{\tt\small \url{https://github.com/LUNAProject22/CausalChaos}}
}

\begin{document}

\maketitle

\begin{abstract}
Causal video question answering (QA) has garnered increasing interest, yet existing datasets often lack depth in causal reasoning. To address this gap, we capitalize on the unique properties of cartoons and construct CausalChaos!, a novel, challenging causal Why-QA dataset built upon the iconic ``Tom and Jerry" cartoon series. Cartoons use the principles of animation that allow animators to create expressive, unambiguous causal relationships between events to form a coherent storyline. Utilizing these properties, along with thought-provoking questions and multi-level answers (answer and detailed causal explanation), our questions involve causal chains that interconnect multiple dynamic interactions between characters and visual scenes. These factors demand models to solve more challenging, yet well-defined causal relationships. We also introduce hard incorrect answer mining, including a causally confusing version that is even more challenging. While models perform well, there is much room for improvement, especially, on open-ended answers. We identify more advanced/explicit causal relationship modeling \& joint modeling of vision and language as the immediate areas for future efforts to focus upon. Along with the other complementary datasets, our new challenging dataset will pave the way for these developments in the field. 
\end{abstract}
\section{Introduction}
Understanding the intricate motivations behind human actions is paramount in developing sophisticated systems capable of nuanced behavior analysis. In real-world scenarios, actions are shaped by a multitude of factors, including personal experiences, emotions, social contexts, and cultural backgrounds. This complexity necessitates a comprehensive approach to unraveling the "why" behind actions, fostering empathy, effective communication, and robust decision-making. Causal video question answering aims to decipher the answers behind characters' actions. Despite the growing interest in causal video QA, existing datasets often fall short, requiring only: 1) \textit{surface-level understanding}; or 2) \textit{involve more of simple word substitution in the QA pairs, rather than causal reasoning} (\eg, ``Q. Why are the cars on the street not moving? A. Parked.").
Recognizing this gap, we embark on the development of a rigorous and challenging causal Video-QA dataset. Our goal is to provide a high-quality resource that rigorously evaluates and advances causal video QA models.

\begin{figure}
\small
    \centering
    \captionsetup{type=figure}
    \begin{subfigure}{0.43\columnwidth}
      \flushright
      \animategraphics[loop,autoplay,poster=1,width=1\columnwidth]{3}{Figs/dataset_intro_frs_indi/5/}{000}{015}
      \label{fig:dataset_eg_a}
    \end{subfigure}
    %
    %
    \begin{subfigure}{0.54\columnwidth}
      \centering
       \raisebox{1mm}[0pt][0pt]{%
\makebox[\textwidth][c]{
\includegraphics[width=\columnwidth]{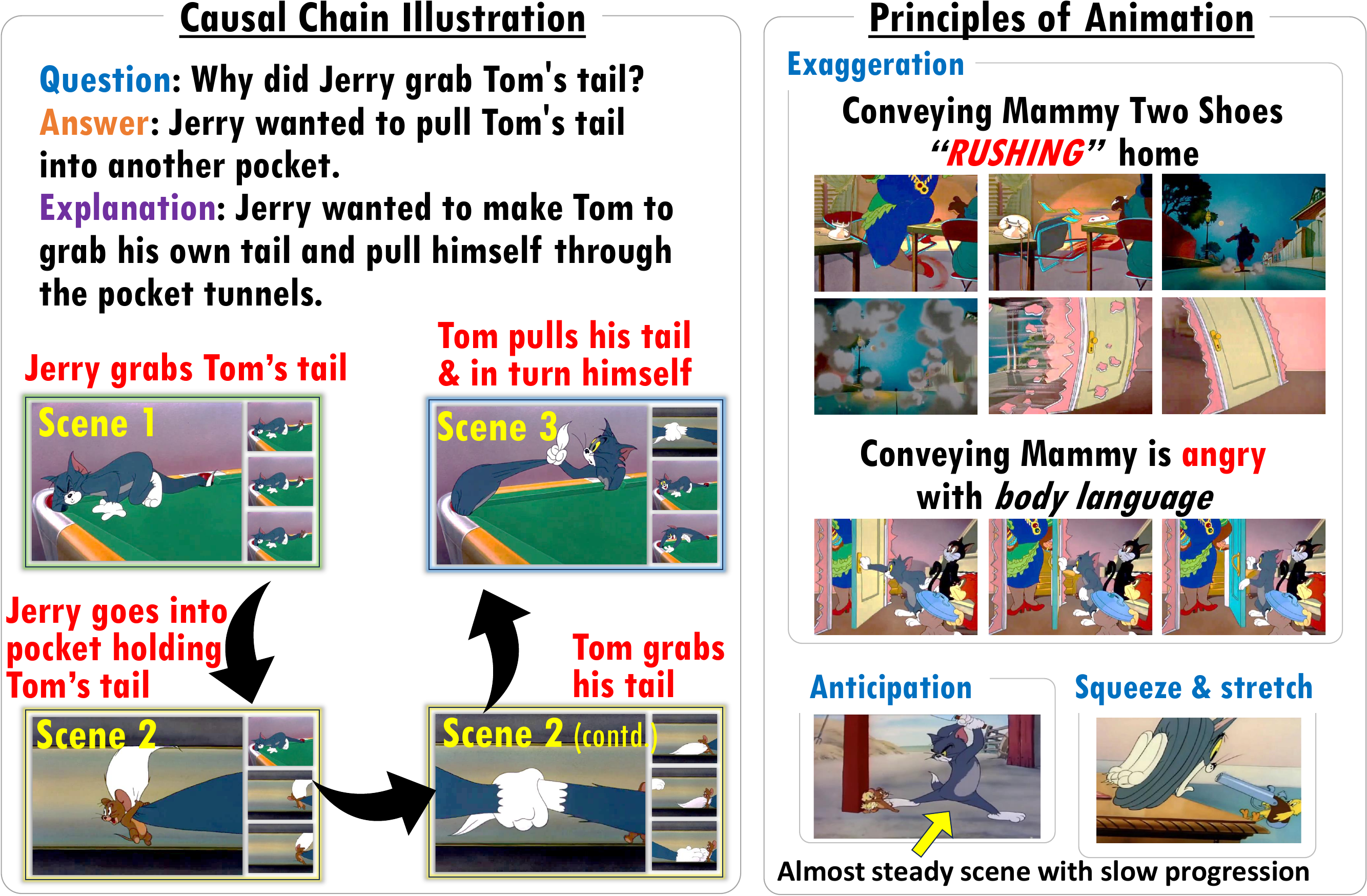}
}}
      \label{fig:dataset_eg_b}
    \end{subfigure}
    \vspace{-0.6cm}
    \caption{\textbf{(Left) Examples of causal questions about characters' actions from our CausalChaos! dataset---based on Tom \& Jerry cartoon series.} Q: question; \textcolor{orange}{A: answer}; \textcolor{violet}{E: explanation}. \textit{Please view in Adobe Reader to play the embedded videos for better explanation.} \textbf{(Middle) Illustration of causal chain, scene changes.} Linking multiple clues/cues embedded in different scenes to solve causal relationships pose a challenge for VideoQA models. \textbf{(Right)} Animators leverage \textbf{Principles of Animation} to \textit{stylize} the \textit{visuals} \& \textit{motions} to \textit{disentangle/highlight key content} of the scene to create \textit{well-defined/unambiguous and effectively communicated cause-and-effect relationships}. The interplay of these factors allow models to focus on solving complex, yet, well-defined, unambiguous causal relationships.}
    \label{fig:dataset_eg}
\end{figure}

\begin{figure}
\begin{subfigure}{0.48\columnwidth}
\centering
\begin{subfigure}{0.5\columnwidth}
  \centering
  \includegraphics[width=\columnwidth]{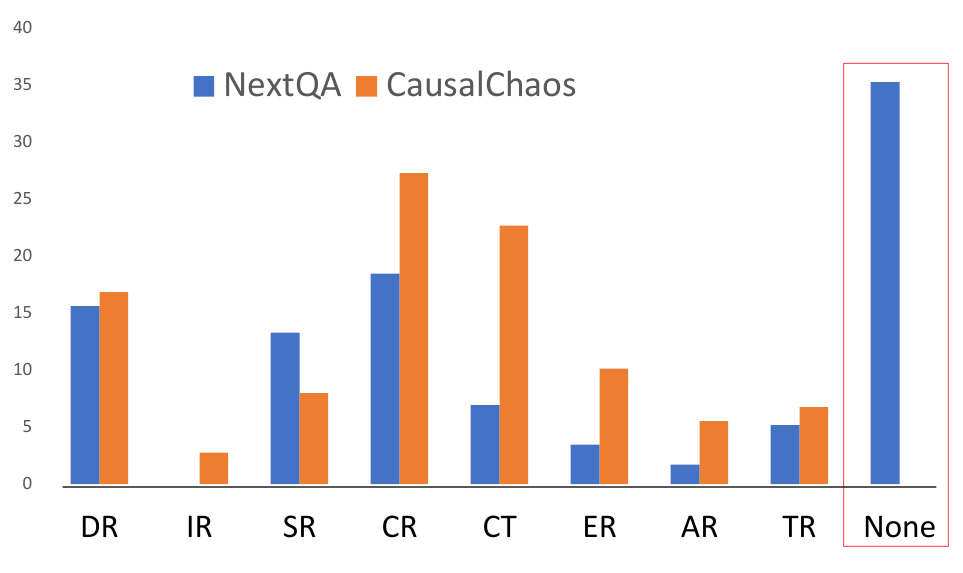}
  \vspace{-0.6cm}
  \caption{}
  \label{fig:reasoning_spectrum_cmp}
\end{subfigure}
\\
\begin{subfigure}{0.9\columnwidth}
\resizebox{\columnwidth}{!}{%
\small
\centering
    \begin{tabular}{@{}lccccc@{}}
    \textbf{Dataset} &
      \textbf{\begin{tabular}[c]{@{}c@{}}MA\end{tabular}} &
      \textbf{\begin{tabular}[c]{@{}c@{}}CCL\end{tabular}} &
      \textbf{\begin{tabular}[c]{@{}c@{}}NOS\end{tabular}} &
      \textbf{\begin{tabular}[c]{@{}c@{}}RS\end{tabular}} &
      \textbf{\begin{tabular}[c]{@{}c@{}}MGA\end{tabular}} \\ \midrule
    NextQA      & \xmark & 1.1 & 1 & narrow & limited \\
    CausalVid  & \xmark & 0.9 & 1 & narrow & limited \\
    IntentQA     & \xmark & 1.3 & 1 & narrow & limited \\
    CausalChaos!        & \cmark & 1.9 & 4 & wide & abundant \\ 
    \end{tabular}}
    \vspace{-0.3cm}
    \caption{{\small{}}}
  \label{fig:causal_chain_histo}
\end{subfigure}
\end{subfigure}
\begin{subfigure}{0.48\columnwidth}
  \centering
  \animategraphics[loop,autoplay,poster=1,width=\columnwidth]{3}{Figs/vvn_frs_pdf/}{000}{015}
  \vspace{-0.5cm}
  \caption{}
  \label{fig:causal_chain_histo}
\end{subfigure}
\caption{\textbf{(a) Types of reasoning demanded by our CausalChaos! dataset.} Reasoning types: DR-\textit{deductive} reasoning; IR-\textit{inductive}; SR-\textit{spatial}; CR-\textit{causal}; CT-\textit{critical thinking}; ER-\textit{emotional}; AR-\textit{abductive}; TR-\textit{temporal}; None-no reasoning required as per the human subjects. None is undesirable, and tend to indicate that questions are less challenging. \textbf{(b) Comparison among CausalChaos! and existing causal videoQA datasets.} 
MA-multilevel answers; CCL-causal chain length; NOS-no.~of~scenes; RS-reasoning spectrum; MGA-multigranular actions. \textbf{(c) Qualitative comparison between CausalChaos! and NextQA (Why-QA) datasets.} CausalChaos! \textcolor{orange}{Answers} and \textcolor{violet}{Explanations} give detailed information regarding the actual cause-and-effect relationships, motivations, emotions covering wide range of reasoning types. Note that, we have temporally cropped videos to retain only the relevant parts from NextQA dataset videos; otherwise, raw videos are longer resulting in unintended problem of temporal localization for models.}
\label{fig:dataset_cmp_quant}
\end{figure}

Drawing on the established benefits of cartoons with cognitive processes among children \cite{schnotz2008functions, shreesha2016does, meringoff1983children, yin2017peer, siong2023use}, we leverage the renowned series \includegraphics[scale=0.20]{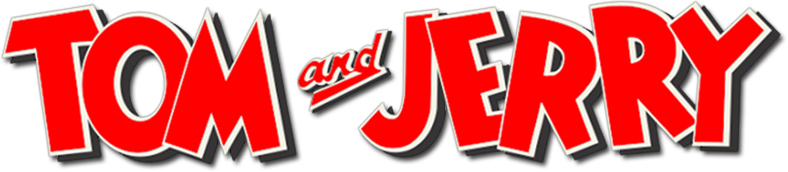} to \textit{create a novel and demanding causal Why-QA dataset} called \textbf{CausalChaos!}. The key characteristics of CausalChaos! are as follows:
\begin{itemize}[leftmargin=*,noitemsep,topsep=0pt]
\item As demonstrated in \autoref{fig:dataset_eg}, \ref{fig:dataset_cmp_quant}, we formulate thought-provoking questions, where the answer (A) and its detailed explanation (E) aim to enhance the model's understanding of the causal chains of visual events in a given video clip.
A causal chain is a sequence of events/actions in which each step influences or leads to the next, creating a cause-and-effect relationship. It illustrates how one event/action produces a subsequent event/outcome.
Compared to existing datasets, CausalChaos! presents longer causal chains, as highlighted in \autoref{fig:dataset_cmp_quant} and discussed in \autoref{sec:related_work}. 

\item Video clips in our dataset feature frequent \textit{scene} and \textit{shot changes}. Here, a scene change or shot change refers to the transition from one visual setting or perspective to another. It typically involves shifting focus to a new location, action, or character within the storyline. 
The \textit{links} and \textit{causes of the causal chains are often dispersed across various scenes}. 
Consequently, models are challenged to exert greater cognitive effort in \textit{connecting multiple events} (scenes) and identifying intermediate causes to comprehend the ``why" questions related to the clips.

\item Despite the complexity and length of the causal chains, they are distinctly \textit{delineated}, \textit{unambiguous}, and effectively \textit{communicated} using \textit{principles of animation} \cite{principles_animations} like staging and exaggeration. This deliberate design allows models to \textit{focus on deciphering causal relationships}.

\item CausalChaos! introduces an added layer of complexity by necessitating the \textit{modeling of actions at various levels of granularity—ranging from sweeping, large-scale movements to nuanced, subtle actions}, such as interpreting emotional cues through facial expressions. 

\item Our dataset demands a \textit{diverse range of reasoning skills}, encompassing deductive, spatial, emotional reasoning, and more, as outlined in \autoref{fig:dataset_cmp_quant} and discussed in \autoref{sec:dataset}.
\end{itemize}

Upon evaluating various state-of-the-art VideoQA models including the recent multimodal instruction tuning models, we found that our dataset remains one of the most challenging causal QA dataset. 
Particularly, we observed that models often: 1) \textit{jumped to conclusions} based on \textit{partial evidence}, rather than considering the full set of evidence; 2) \textit{failed to engage in true causal reasoning}, opting instead for shortcuts like object/action-noun/verb matching to select answers.
Based on this, we identified \textit{more advanced/explicit causal relationship modeling and jointly modeling vision and language} as the immediate areas for future efforts to focus upon. Further, we show that similar to how cartoons help children better connect cause and effects, they can help VideoQA models as well. We \textit{incorporated our dataset with the NextQA}~\cite{nextqa}, a \textit{real-world dataset} and found that it \textit{brings some improvements in why questions}. What is more, \textit{incorporating our dataset} brings improvement on \textit{non-Why questions} as well. 

Overall, we believe our CausalChaos! dataset presents challenges spanning the entire VideoQA pipeline, from \textit{deciphering intricate videos} to \textit{processing complex questions} and \textit{discerning nuanced answers}, stimulating research in \textit{video processing}, \textit{causal reasoning}, \textit{language modeling}, and \textit{joint modeling}.
Along with the other complementary datasets, such as \cite{nextqa, causalvidqa, intentqa, videochat2, wu_star}, our new challenging dataset will pave the way for these developments in the field. 

 \section{Related Work}
\label{sec:related_work}
To drive the progress in VideoQA, researchers have developed various datasets \cite{moviefib, msvd, youtube2text, openendedqa_zhou, zeng2017leveraging, Videocontextqa, stvqa, stvqa2, activitynetqa, socialiq, agqa, wu_star} with distinct focuses and contributions---see the survey~\cite{zhong2022video}. In the following, we mainly compare our work with prior art focusing on causal QA literature. We provide a detailed discussion on various VideoQA datasets and models in the \supplementary. Our work is inspired from a few datasets: {\small{{\textit{CLEVRER}}}} \cite{clevrer, clevrerhumans}, {\small{\textit{NextQA}}} \cite{nextqa}, {\small{\textit{CausalVidQA}}} \cite{causalvidqa}, {\small{\textit{IntentQA}}} \cite{intentqa} that cover causal reasoning. However, they have limitations that our dataset aims to fill: 
\begin{itemize}[leftmargin=*, noitemsep, topsep=0pt]
    \item \textit{Limitation in scope}: \eg, \cite{clevrer, clevrerhumans} only consider \textit{collision events of simple inanimate objects} such as cuboids; as a result, \textit{lack: actors}, their \textit{characteristics} such as \textit{emotions}, \textit{intentions}. Our dataset has actors whose actions are shaped by \textit{emotions}, \textit{intentions}, \textit{context}, etc.

    \item \textit{Lack of precise temporal annotations} (e.g., \cite{nextqa, causalvidqa}) \textit{inadvertently} involves \textit{temporal localization} \& may create a \textit{false sense of difficulty}, while the questions may \textit{not be challenging}. We alleviate this problem by providing precise temporal annotations and focusing on designing \textit{truly complex} and challenging QA. 
    
    \item \textit{Lack of complexity in questions} (\eg \cite{nextqa, causalvidqa, intentqa}). Based on human studies, we found that a majority of causal QA in these datasets were flagged as \textit{not requiring any notable reasoning} (\autoref{fig:dataset_cmp_quant}\textcolor{red}{(a)}).  We also attempted to \textit{quantify} the \textit{complexity in reasoning} by computing the \textit{lengths of causal chains} involved in their QAs vs. ours using GPT-4o \cite{gpt4}. Causal chains in \cite{nextqa, causalvidqa, intentqa} are shorter (\autoref{fig:dataset_cmp_quant}\textcolor{red}{(b)}. In comparison, our dataset has \textit{longer causal chains}---average length 1.9---posing a challenge for models in \textit{connecting multiple events} or cues. Our dataset also demands \textit{wide spectrum of types of reasoning} (\autoref{fig:reasoning_spectrum_cmp}).

    \item Scenes in \cite{clevrer, clevrerhumans, nextqa, causalvidqa, intentqa} are \textit{less dynamic}, mostly involving a \textit{single scene}. In comparison, our dataset averages about \textit{four scenes}. \textit{Rapid scene changes} \& \textit{dynamic interactions} challenge VideoQA models to link \textit{context} \& \textit{cues across different scenes}, modeling cause-and-effect relationships over \textit{longer causal chains} (more details in \autoref{sec:uniqueness_dataset}). While similar challenges have been acknowledged in other computer vision problems \cite{pavlakos2022human}, they remain unexplored in VideoQA.

    \item \textit{Lack of hard negatives.} Current datasets may not emphasize on including hard incorrect options. Due to this, models may not be required to do causal reasoning, but rather they can exploit \textit{shortcuts like object/action-noun/verb matching} in vision-language spaces to select correct answer. To address this limitation, we develop \textit{hard negative selection strategies}.

    \item \textit{No hierarchical answers with different level of explanation.} These datasets contain a \textit{single-level answer to `Why'}-questions. On the other hand, we richly annotate our dataset to provide \textit{two-level answers to `Why'-questions}---1) direct/immediate cause; 2) deeper explanation. Quality of answers is further enhanced, and are \textit{more informative} as a consequence of the \textit{more complex questions}.
\end{itemize}
\section{CausalChaos! Dataset}
\label{sec:dataset}
This section details the construction of our VideoQA dataset designed to challenge causal reasoning, covering its \textit{video source}, \textit{annotation process}, \& \textit{quality checks} to ensure high-quality annotations. We then discuss the \textit{unique attributes} that make our dataset valuable for causal VideoQA tasks.

\subsection{Dataset Construction}
\paragraph{Video source.} 
To focus on \textit{visual reasoning}, we selected the timeless \textit{silent} cartoon series, ``\textit{Tom \& Jerry}" (1940)\footnote{We only provide the annotations; Videos can be obtained from: https://www.warnerbros.com/tv/tom-and-jerry-1965-volume-1.}, spanning over six seasons and 161 episodes. Silent cartoons, which lack other modalities like dialogue, foster visual reasoning in children. Similarly, we hypothesize that using silent cartoons for VideoQA will enhance visual reasoning. The Tom \& Jerry series is ideal for a causal reasoning dataset, offering \textit{abundant segments with diverse cause-and-effect relationships}. Concurrent work \cite{li2024image} on causal \textit{image} generation also leverages Tom \& Jerry.

\paragraph{Annotations.} 
Each dataset sample includes the following annotations: \{\textit{start frame}, \textit{end frame}, \textit{Question}, \textit{Answer}, \textit{Explanation}\}. Details in the following.\\
\noindent \textcolor{black}{$\bullet$} \textbf{Questions} are crafted to capture the \textit{why} or reasoning behind the \textit{actions} of characters in Tom \& Jerry cartoons. To cultivate comprehensive \& deeper video understanding capabilities in models, we formulated \textit{thought-provoking} causal questions where cause \& effect are connected by \textit{longer causal chains}. Questions also extend \textit{beyond explicit} visual cues, encompassing gestures \& expressions, to delve into the \textit{characters' underlying intentions} \& \textit{goals}. 
Annotators generated questions while watching the video for the first time to mimic how models assess unseen clips. They then rewatched the videos multiple times to create more critical thinking questions. To focus on visual reasoning, annotators watched the videos without audio, ensuring no audio cues were included in the dataset.\\
\noindent \textcolor{black}{$\bullet$} \textbf{Multi-level Answers}.
The first-level or the \textcolor{orange}{Primary answer}, represents a \textit{literal} or \textit{direct} cause or form of response. It is accompanied by a deeper form of \textcolor{violet}{Explanation}, which considers the \textit{broader context} of the scene, the \textit{thoughts}, \textit{feelings}, \textit{intentions} of characters, \& their actions. This deeper explanation also takes into account potential consequences \& provides further reasoning to support the primary answer. It includes reasons \& additional details to comprehensively address the question. Examples shown in \autoref{fig:dataset_eg} \& \supplementary.\\
\noindent \textcolor{black}{$\bullet$} \textbf{Temporal annotations}.
For each QAE set, the \textit{Start} and \textit{End frame numbers} are recorded to cover the entire scene, including contextual frames that support the reasons behind the actions.

\paragraph{Quality Check.}
To ensure dataset quality and reliability, we implemented a two-stage quality check process. First, quality checkers assessed episodes they did not annotate. Then, they reviewed episodes they did not check in the first stage. Multiple quality checkers carefully reviewed and verified curated question and answer sets for logical fallacies, timestamp inconsistencies, grammatical errors, and spelling mistakes. They also excluded annotations containing audio or text aspects that the model cannot comprehend. To prevent direct overwriting, edits were flagged in a different font color for further discussion. Flagged inconsistencies or errors were resolved through discussions and consensus among checkers and annotators. The team referred to episode synopses from Tom and Jerry Wiki \cite{tom_and_jerry_wiki}, an online encyclopedia, as a third-party opinion and for fact-checking.

\paragraph{Dataset Statistics.}
Annotators reviewed all 161 episodes across 6 seasons, generating 4,945 detailed \textcolor{blue}{Question}-\textcolor{orange}{Answer}-\textcolor{violet}{Explanation} sets. This comprehensive dataset covers various scenes, characters, and events, providing a solid foundation for understanding and answering questions about the series. Additional statistics and word clouds for answers and explanations are available in the \supplementary.

\begin{figure}
    \centering
    \includegraphics[width=\columnwidth]{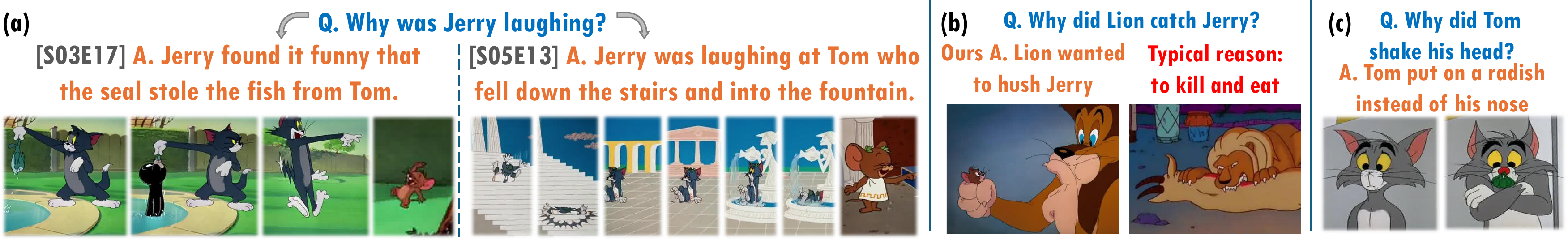}
    \vspace{-0.6cm}
    \caption{\textbf{Grounded in diverse visual information.}}
    \label{fig:visual_grounding}
\end{figure}
\begin{figure}
    \centering
    \includegraphics[width=\columnwidth]{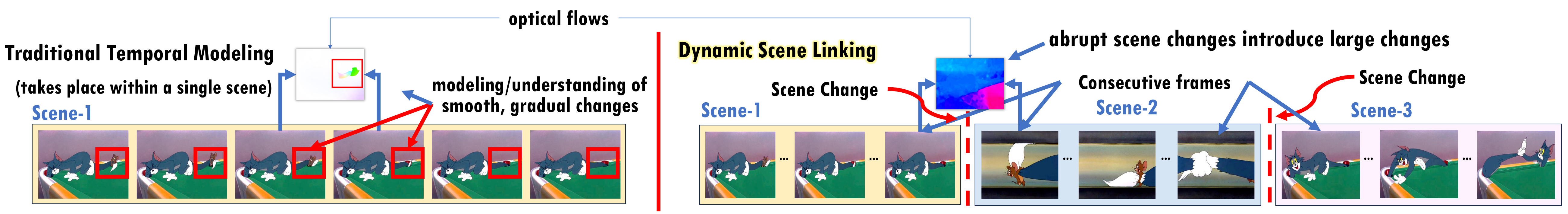}
    \caption{\textbf{Traditional temporal modeling vs Dynamic scene linking.} Notice the abrupt scene change, which causes disruption in visual flow, resulting in large amplitude and widespread optical flow.}
    \label{fig:dynamic_scene_linking}
\end{figure}
\subsection{Unique Characteristics of our CausalChaos! Dataset}
\label{sec:uniqueness_dataset}
In the following, we delve into the untapped unique properties of our dataset, which both \textit{complement existing video understanding datasets} and present \textit{novel challenges to VideoQA}, particularly from a \textit{causal reasoning perspective}.

\begin{enumerate}[wide, labelwidth=!, labelindent=0pt,topsep=0pt]
\item \textbf{Multi-level answers} offer a \textit{richer and more nuanced perspective}, which leads to \textit{deeper insights}---considering not only the \textit{immediate circumstances} but also underlying \textit{psychological}, \textit{social}, and \textit{personal} factors that contributed to \textit{characters' decisions}. By employing \textit{multi-level answers alongside thought-provoking questions}, we can cultivate \textit{deeper reasoning} and \textit{analysis of characters' actions}, enabling the training and evaluation of models on complex question-answer pairs that demand a \textit{comprehensive understanding} of \textit{longer causal chains}. Further elaborated discussion on the importance of multi-level answers to `Why'-questions included in \supplementary.

\item \textbf{Grounded in diverse \textit{Visual} \& \textit{Motion} information.} The Tom \& Jerry cartoon series is \textit{visually rich} and \textit{dynamic}, with \textit{intricate scenes}, \textit{actions}, and \textit{character interactions varying across episodes}. We had annotators craft questions where the \textit{answer is grounded in the video}. This grounding compels the \textit{model to analyze the video} for a \textit{broader range of details} and \textit{clues} to provide \textit{meaningful} answers.
For example, consider \autoref{fig:visual_grounding}\textcolor{red}{(a)}, same question appears in different episodes, but their answers/context are completely different. This demonstrates that it is crucial for VideoQA models to understand visual information to answer correctly on our dataset.
Multimodal nature of VideoQA task is enhanced by the presence of \textit{unusual situations} in cartoons---typically not found in real-world datasets as shown in \autoref{fig:visual_grounding}\textcolor{red}{(b,c)}. 
Since the answers are embedded in videos, models cannot rely on \textit{correlations/biases} in their training data; they must thoroughly process and understand the content to answer correctly. 
This is crucial even with large language models (LLMs) and embeddings, which are also susceptible to \textit{frequency-based biases} and hallucinations \cite{li2024dawn, liu2024survey, zhou2023explore}. For example, \textit{Lion} is often associated with \textit{killing} in LLM embeddings. In a qualitative analysis, we found that models might choose an answer involving '\textit{killing}' Jerry as the \textit{motive} for the Lion-character's action, while the Lion-character was actually trying to \textit{quiet} Jerry \autoref{fig:visual_grounding}\textcolor{red}{(b-left)}. 
Revealing such biases has been considered in other computer vision problems \cite{barbu2019objectnet, he2016human}, but is yet to be introduced in VideoQA. \textit{Our dataset offers a valuable opportunity to address this challenge}. Additionally, understanding Tom \& Jerry cartoons requires modeling of \textit{details} and \textit{actions at varying levels of granularity}---from \textit{sweeping movements} to \textit{subtle emotional cues} via facial expressions.

\item \textbf{Focus on causal reasoning in visually dynamic scenes.}
Our dataset focuses on causal reasoning in Tom and Jerry cartoons, characterized by dynamic and rapidly changing scenes. Characters and objects may appear, disappear, and reappear, limiting the model's access to partial observations. 
Video understanding models struggle to track and understand context. To answer causal questions, these models must link events to form a causal chain despite rapid scene changes and dynamic interactions.
We term this Dynamic Scene Linking (DSL), distinct from traditional temporal modeling, which typically focuses on the gradual transitions within a scene as shown in \autoref{fig:dynamic_scene_linking} (more details in \supplementary). Related challenges are noted in other computer vision problems \cite{pavlakos2022human, wong2015data} but not yet in VideoQA. We hypothesize that humans understand cartoons by forming a mental ``world model" of the scene, which helps bridge gaps between discontinuous scenes and provides a coherent overall context. 
Current VideoQA datasets often lack coverage of such complex scenarios, unlike our dataset.

\item \textbf{More challenging and requiring cognitive effort.}
We used GPT-4o \cite{gpt4o} to compute the lengths of causal chains of \textit{100 randomly chosen QAE pairs} from \textit{our dataset} and \textit{existing causal(-Why) VideoQA datasets} \cite{nextqa, causalvidqa, intentqa}. Results are presented in \autoref{fig:dataset_cmp_quant}. We observed that the QAE pairs in \textit{our dataset have longer causal chains than existing QA datasets}, potentially suggesting our QAs are more complex.
Identifying and solving \textit{longer causal chains} in the process of answering causal (why) questions involve linking together multiple cues/clues, which requires significant cognitive efforts on the part of VideoQA models, especially, when the clues/cues are embedded in different scenes (described above), as in our dataset.

\item \textbf{Leveraging principles of animation} \cite{principles_animations} (shown in \autoref{fig:dataset_eg}) such as \textit{timing}, \textit{squash and stretch}, \textit{anticipation}, \textit{staging}, and \textit{exaggeration} aid in: 1) ``\textit{highlighting}" \textit{key movements}, \textit{emotions}, \& \textit{storytelling}; 2) Consequently, greatly aid in \textit{establishing clear cause-and-effect relationships} and \textit{effectively communicating them}. These principles can be thought of as \textit{spatiotemporal} counterparts of \textit{caricatures}, which exaggerate and manipulate facial and bodily features, and these have been shown to improve facial recognition rates \cite{rhodes1987, han2023caricaturing, mauro1992caricature}. In other domains of computer vision, some works \cite{sun2023alpha, xu2023side, xu2023open, ding2023open} acknowledge the advantages and devise methods to \textit{disentangle the area of interest}, enabling more \textit{targeted} processing and \textit{comprehension of image contents}. Similarly, we hypothesize that the \textit{principles of animation} can provide ``\textit{hints/guidance}", but the models still need to be able to \textit{leverage this guidance to solve causal relationships}. Overall, \textit{causal relationships in our dataset are complex, longer}, but at the same time they are \textit{unambiguous/well-defined} using principles of animation, allowing models to \textit{focus on deciphering causal relationships}.

\begin{figure*}[!t]
    \centering 
    \includegraphics[width=\linewidth]{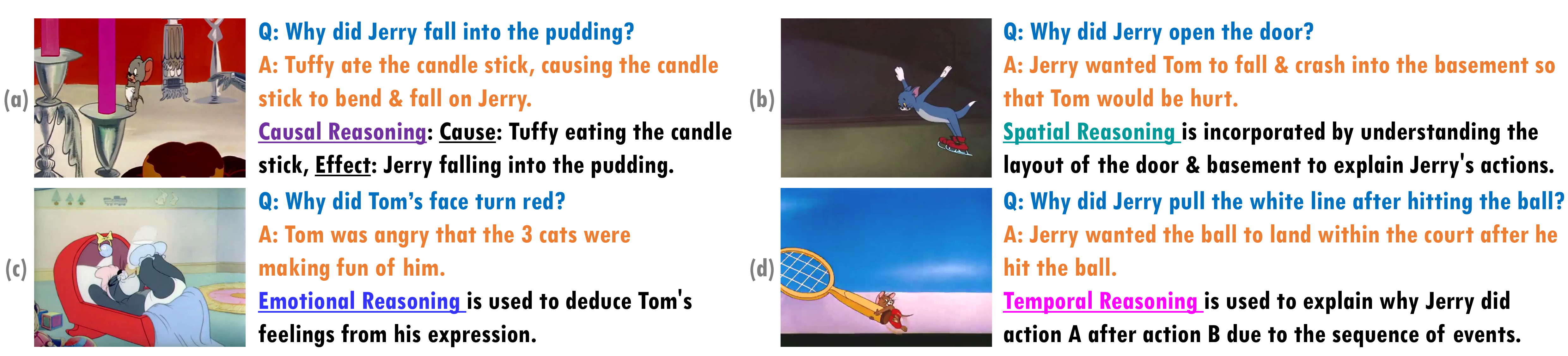}
    \caption{\textbf{Examples of various types of reasoning required by our dataset.} \textit{Please zoom in \& view in AdobeReader to play the embedded videos.} Further types of reasoning visualized in \supplementary.}
    \label{fig:dataset_reason_viz}
\end{figure*}
%
%
%

\item \textbf{Wide spectrum of types of reasoning required.} Our dataset demands various types of reasoning like 1) \textit{Deductive}; 2) \textit{Inductive}; 3) \textit{Spatial}; 4) \textit{Causal}; 5) \textit{Critical thinking}; 6) \textit{Emotion}; 7) \textit{Abductive}; and 8) \textit{Temporal}. We have visualized examples demonstrating these reasoning types in \autoref{fig:dataset_reason_viz}. It is also possible that a question may involve more than one type of reasoning. We conducted \textit{human studies} to determine the types of reasoning required by our dataset and NextQA and \textit{observed that our dataset demands a wider range of reasoning than NextQA} (results in \autoref{fig:reasoning_spectrum_cmp}). More information and definitions of types of reasoning and human studies are provided in \supplementary. 
\end{enumerate}

\subsection{Benchmarks}

\paragraph{Question-types/Tasks.} Our dataset comprises two types of questions.  
The first is Multiple Choice Question Answering (\textbf{MCQA}). VideoQA models are provided with the question, the associated video footage, the correct answer, and four incorrect answers; the task for the models is to pick the correct answer. The second is Open-ended Answer Generation (\textbf{OEAG}). The models are provided only with the question and the video footage, and the models have to generate the answers in natural language.

\paragraph{Incorrect answer/option mining.} 
In designing MCQA, answer options should be distinct from the correct answer \& each other but semantically similar enough to challenge reasoning beyond commonsense. This prevents models from taking shortcuts like simple object/action matching in vision-language spaces. Using randomly chosen answers as incorrect options is unproductive and makes the correct answer too obvious, so we avoid this strategy. Instead, we introduce these strategies:

\vspace{-0.2cm}
\begin{enumerate}[wide, labelwidth=!, labelindent=0pt, noitemsep,topsep=1pt]
    \item \textbf{Vanilla Hard Negative mining.} A pre-trained Sentence-Bert measures semantic similarity. For each question, we sample the top-10 other questions based on cosine similarity of Sentence-Bert embeddings. From these, we sample associated answers and again apply cosine similarity between the correct and candidate answers. Using the equation $\alpha cossim(Q_i,Q_n) \cdot \beta cossim(A_i,A_n)$, with adjustable weights $\alpha$ and $\beta$, we select the top 4 answer candidates. To avoid overly similar answers, we penalize high similarity scores by setting $\beta$ to 0.001, reducing the candidate's overall score.
    
    \item \textbf{Causal-Confusion Negative generation.} 
    Causal relationship modeling is crucial for causal VideoQA, but maybe given less emphasis in existing datasets/benchmarks. We hypothesize that, relationships might not be adequately modeled by existing models, instead models may be selecting answers based on basic object/actor matching in vision-language spaces. To test our hypothesis \& underscore the relationship modeling, we introduce hard negatives in which the objects/actors remain unchanged, but the relationships between them are altered or alternate, but plausible scenarios are created. For instance, the effect is inverted (\eg``Tom was hitting Jerry"$\rightarrow$``Tom was \textit{not} hitting Jerry") or the causal agents are swapped (\eg``\textit{Tom} was hitting \textit{Jerry}"$\rightarrow$``\textit{Jerry} was hitting \textit{Tom}"). We leverage LLMs to generate causally-incorrect options. Examples can be found in the \supplementary.    
    
\end{enumerate}

\paragraph{Design of Train-Val-Test splits.}
We typically partition the dataset into 70\% for training, and 15\% each for validation and testing. Additionally, we employ specific strategies for training and testing.
\vspace{-0.2cm}
\begin{enumerate}[wide, labelwidth=!, labelindent=0pt, noitemsep, topsep=0pt]
    \item \textbf{Uniformly Distributed \textit{Seen}-episode testing (UD).}  
    Dataset samples are randomly divided into train-val-test splits without constraints on which episodes or parts are included in each split. Consequently, uniformly distributed chunks of all episodes from all seasons are likely seen during training, potentially giving VideoQA models sparse storylines.
    \item \textbf{Consecutive Partially \textit{Seen}-episode testing (PS).} 
    The train set is drawn from the first 70\% of each episode, the val set from the middle 70-85\%, \& the test set from the last 15\%. This setup allows the model to have partially "seen" episodes during training \& make "educated guesses" during testing.
    \item \textbf{\textit{Unseen}-episode testing (UN).} In this case, a certain 70\% of the episodes are reserved for train set; 15\% of the remaining episodes are reserved for val set; and the remaining 15\% of the episodes are used to prepare the test set. So, testing is done on entirely novel, unseen episodes and storylines.
\end{enumerate}
\section{Experiments}
\label{sec:experiments}
We benchmark SOTA VideoQA methods on our new causal action QA dataset for two tasks: 1) \textbf{MCQA} and 2) \textbf{OEAG}. Then, we explore if our dataset aids real-world QA.

\subsection{Benchmarking on MCQA}
\paragraph{Baselines.} 
Following prior work \cite{nextqa}, we benchmark the performance of these models on our causal action QA dataset: \textit{BlindQA} \cite{evqa}, \textit{EVQA} \cite{evqa}, \textit{CoMem} \cite{comem}, \textit{HME} \cite{hme}, \textit{HCRN} \cite{hcrn}, and \textit{HGA} \cite{hga}. We also evaluate recent vision-language models like \textit{MIST} \cite{mist}, \textit{BLIP-2} \cite{blip2}, and multimodal instruction tuning models \textit{Video-LLaMA} \cite{videollama} and \textit{VideoChat2} \cite{videochat2}, \textit{GPT-4o} \cite{gpt4o} which excel on vision-language tasks. Further details on each model are provided in \supplementary.

\paragraph{Evaluation Protocols.} Models are evaluated under these protocols, with \textbf{Accuracy} as the metric:
\vspace{-0.2cm}
\begin{itemize}[wide, labelwidth=!, labelindent=0pt, noitemsep,topsep=0pt]
    \item \textbf{Protocol 1.} Models' ability to select the \textit{correct \textcolor{orange}{Answer (\textbf{A})}} is only evaluated. Their ability to select the correct explanation is not taken into consideration. The chance accuracy in this protocol is $1/5$. Results from this protocol are presented under columns marked as \textbf{A} in \autoref{tab:res_mcq_reg}.
    \item \textbf{Protocol 2.} Models' ability to select the \textit{correct \textcolor{orange}{Answer (\textbf{A})} as well as the correct \textcolor{violet}{Explanation (\textbf{E})}} is evaluated. If the model can select the correct answer, but not the correct explanation, then the model is considered to have failed. In this protocol, the chance accuracy becomes $1/5 \times 1/5 = 1/25$. Results from this protocol are presented under columns marked as \textbf{A+E} in \autoref{tab:res_mcq_reg}.
\end{itemize}

\begin{wraptable}[16]{R}{0.45\columnwidth}
\vspace{-.35cm}
\resizebox{\columnwidth}{!}{%
\small
\centering
\setlength\tabcolsep{5pt}
\begin{tabular}{@{}lcccccc@{}}
\toprule 
 & \multicolumn{2}{c}{\textbf{UD}} & \multicolumn{2}{c}{\textbf{PS}} & \multicolumn{2}{c}{\textbf{UN}} \\ \cmidrule(l){2-3} 
 \cmidrule(l){4-5} 
 \cmidrule(l){6-7} 
\multirow{-2}{*}{\textbf{Model}}              & \textbf{A} & \textbf{A+E} & \textbf{A} & \textbf{A+E} & \textbf{A} & \textbf{A+E}                 \\ 
\cmidrule(r){1-1} \cmidrule(lr){2-2} \cmidrule(lr){3-3} \cmidrule(lr){4-4} \cmidrule(lr){5-5} \cmidrule(lr){6-6} \cmidrule(l){7-7}
Chance    & 20.00      & 04.00        & 20.00      & 04.00        & 20.00     & 04.00 \\
\textbf{BlindQA}    & 29.38      & 13.07        & 26.51      & 11.54        & 25.13      & 11.02 \\
\textbf{EVQA}             & 29.38      & 13.48        & 31.32      & 13.32        & 27.82      & 14.65                        \\
\textbf{CoMem}             & 32.08      & 13.88        & 26.10      & 09.89        & 23.12      & 09.27                        \\
\textbf{HME}              & 32.35      & 14.02        & 29.53      & 12.36        & 25.13      & 10.22                        \\
\textbf{HCRN}             & 32.48      & 16.98        & 32.01      & 14.97        & 25.67      & 12.23                        \\
\textbf{HGA}               & 31.40      & 15.36        & 29.40      & 13.19        & 28.23      & 13.84                        \\
\textbf{MIST}             & 62.22      & 44.88        & 62.22      & 42.86       & 55.91      & 37.90                       \\
\textbf{MIST-CC\textsuperscript{\textdagger}}     & \underline{63.34}      &  \underline{46.80}     & \underline{62.50}      & \underline{43.54}      & \underline{56.18}     &  \underline{39.25}                    \\
\multicolumn{7}{c}{\textit{Zero-shot}} \\ 
\midrule
\textbf{BLIP-2}  &  43.67     &  23.32     & 45.88      & 24.18      &  46.64     & 26.48                               \\
\textbf{Video-LLaMA}  &  35.00     &  11.73     & 35.16      & 9.34      & 29.62      & 9.68                       \\
\textbf{VideoChat2}  &  38.14     &  15.36     & 38.91      & 15.93      & 40.83      & 15.99                      \\
\textbf{GPT-4o}  & \textbf{63.64}      & \textbf{48.17}      &  \textbf{63.84}     & \textbf{49.79}      &  \textbf{64.82}    &  \textbf{52.23}                    \\
\bottomrule
\end{tabular}
}
\vspace{-1em}
\caption{\textbf{MCQA Results on our dataset.}\textsuperscript{\textdagger}We design multitask version of MIST that learns to generate causal chains as an auxiliary task.} 
\label{tab:res_mcq_reg}
\end{wraptable} 

\paragraph{Quantitative results.} The performances of the baseline models for both protocols are presented in \autoref{tab:res_mcq_reg}. In general, we observe that the performances of most of the baseline approaches are low. Most of the models gain only around 11-12\% improvement over the chance accuracy for both protocols. MIST seems to be doing exceptionally well compared to other models. We hypothesize that this could be because MIST being geared for long-form video understanding, is better able to handle long video contexts, which are frequent in our dataset. Overall, we observed the following failure modes for models: \textbf{1)} \textit{Limited Evidence Consideration}: Models often focus on a small subset of evidence, neglecting other relevant cues distributed throughout the sequence, which is problematic for datasets with long causal chains like ours. \textbf{2)} Models failed because they tried to exploit \textit{shortcuts} like object-noun or action-verb matching in video-language spaces instead of focusing on causal relationship modeling---such \textit{shortcuts can lead to wrong predictions on our dataset due to our hard negative mining}, correct \& incorrect answers would likely contain these nouns/verbs. The most effective way to discriminate on our dataset is by inferring the causal relationship. We have discussed further shortcomings with \textbf{qualitative results} in \supplementary. To mitigate these shortcomings, we design a MIST-based multitask model, \textbf{MIST-CC}, which, in addition to doing VideoQA, also learns to generate the causal chain from Video-Question features (groundtruth causal chains are generated by GPT-4 from QAE pairs). We found that this simple modification mildly boosts the performance. We have provided further details on MIST-CC in \supplementary. Moving to VLM/MLLMs, we observe that they perform better than traditional models (\cite{evqa, hga, hcrn}), except MIST. However, GPT-4o released only in late May 2024, outperformed all the models. Although, it is closed source model, we hypothesize its vision and language capabilities maybe significantly better than other VLMs. It was able to incorporate small details like facial expressions. To glean insights into it, we also asked it to give us its reasoning, and found that it analyzes each option individually and then selects the most likely answer. This is close to (or at least mimicking) how humans would approach this task. We believe this, at least on surface, seems to be going beyond just correlation-based answer picking as in non-VLM/MLLMs.

\textbf{Comparing the two protocols}, we find that Protocol 2, where the model has to select both the correct answer and the correct explanation is significantly more difficult than selecting just the correct answer for all the models.
\textbf{Comparing splits}, 
We observe that UD is the easiest split across all models, followed by PS, with UN being the most difficult. This intuitive order suggests that understanding past events or storylines may aid in reasoning about current events or actions.

\begin{wraptable}[6]{R}{0.35\columnwidth}
\vspace{-.35cm}
\resizebox{\columnwidth}{!}{%
\small
\centering
\setlength\tabcolsep{1pt}
\begin{tabular}{@{}lccc@{}}
    \toprule
    \textbf{Negative type} & \textbf{MIST} & \textbf{MIST-CC} & \textbf{GPT-4o}\\
    \cmidrule(r){1-1} \cmidrule(lr){2-2} \cmidrule(lr){3-3} \cmidrule(l){4-4}
        Vanilla Hard  & 62.22 & 63.34 & 63.64\\
        Causal-Confusion  & \underline{55.80} & \underline{58.76} & \underline{54.95}\\
        \bottomrule
    \end{tabular}
    }
    \vspace{-0.2cm}
    \caption{\textbf{Impact of causal-confusion.}}
    \label{tab:res_causal_confusion}
\end{wraptable}

We further conducted an experiment where we tested the best performing models, MIST, MIST-CC, \& GPT-4o, on our \textbf{Causal-Confusion} set, where incorrect answer choices have the same objects and actors as the correct answer option, but cause-and-effect relationships are reversed or altered. 
We observed a \textit{significant drop} in performance of all models as shown in \autoref{tab:res_causal_confusion}. 
This could potentially be due to \textit{causal relationships being not modeled adequately} even by such advanced VideoQA models. Performance of MIST-CC dropped relatively less, potentially because the auxiliary task of generating causal chains may have enhanced the understanding of causal relationships. Interestingly, we also found the S-BERT similarity score to be above 90\% for an action (\eg, Tom is running after Jerry) and its Causal-Confusion version (\eg, Tom is \textit{not} running after Jerry); while these sentences would be opposite/different in terms of human perception. We believe that \textit{Causality might be more overlooked than we think in various fields, not just in computer vision}.

\subsection{Benchmarking on Open-ended Answer Generation (OEAG)}

\paragraph{CapsMIX Performance metric.} We measure the performance of OEAG in terms of BLEU-1,2,3 \cite{bleu}, METEOR \cite{meteor}, ROUGE \cite{rouge}, SPICE \cite{spice}, CIDEr \cite{cider} \& Sentence-BERT \cite{sentencebert} scores by comparing with the ground truth answers as done by captioning \& QA literature. The wide range of metrics complicates model comparison, so we introduce Caps-MIX (Captioning Metrics Integration eXpert), which normalizes and integrates all scores into a single metric, simplifying comparisons and combining the unique strengths of individual metrics.

\begin{wraptable}[12]{R}{0.2\columnwidth}
\vspace{-.4cm}
\resizebox{\columnwidth}{!}{%
\small
\centering
\setlength\tabcolsep{5pt}
\begin{tabular}{@{}lc@{}}
\toprule
\textbf{Model}
 &
\textbf{CapsMIX}
   \\
   \cmidrule(r){1-1} \cmidrule(l){2-2}

\textbf{BlindQA} &
  2.7646 \\

\textbf{UATT} &
  3.3928 \\
\textbf{HME} &
  3.0975 \\
\textbf{HGA}&
  3.7872 \\
  
  \textbf{BlindGPT-2} &
  6.6006 \\
  
\textbf{VisionGPT-2} &
  \underline{6.7582} \\
  
  \midrule

\textbf{BLIP-2} &
  1.8931 \\

  \textbf{Video-LLaMA} &
  2.4464 \\

  \textbf{VideoChat2} &
  \textbf{3.9524} \\ 
  \textbf{GPT-4o} &
  2.9851 \\ \bottomrule\\
\end{tabular}%
}
\vspace{-0.5cm}
\caption{\textbf{OEAG results on our dataset (UD split).}}
\label{tab:res_oe_consolidated}
\end{wraptable}
\paragraph{Baselines.} Following prior work \cite{nextqa}, we benchmarked the performance of \textit{EVQA} \cite{evqa}; \textit{UATT} \cite{uatt}; \textit{HME} \cite{hme}; \textit{HGA} \cite{hga} on OEAG task. We also report the zeroshot performance of recent multimodal video understanding models \textit{BLIP-2} \cite{blip2}, \textit{Video-LLaMA} \cite{videollama}, \textit{VideoChat2} \cite{videochat2}, \textit{GPT-4o} \cite{gpt4o}.

We evaluated the baselines for generating answers and explanations. The performances of various models are summarized in \autoref{tab:res_oe_consolidated}. For full results, see \supplementary.
Overall, we observe that models struggle significantly with open-ended generation, including some recent VLMs and MLLMs. Despite GPT-4o's strong performance on MCQA, it also falters on OEAG. These models likely perform better on MCQA by eliminating incorrect choices, but they fail to genuinely understand videos and perform causal reasoning for standalone answer generation. VideoChat2 seem to be doing exceptionally well, perhaps, because it was specifically trained on various video understanding tasks and datasets, including causal reasoning task. 
Comparing split-type wise, we found that as in the case of MCQA, UD was the easiest split. 

Inspired by the success of LLMs, we experimented with leveraging GPT-2 \cite{gpt2}, a publicly available LLM, as our natural language answer generator. We found that a pre-trained version out of the box did not work well---it generated mostly random, unrelated words. However, upon simple training on the train set, it performed significantly better than the baselines we considered as shown in \autoref{tab:res_oe_consolidated} (BlindGPT-2). However, it is unclear if these LLMs have ``seen" Tom and Jerry scripts during their pretraining stage. If so, then pretraining on relevant scripts followed by finetuning on our dataset could potentially be a reason for GPT-2's good performance. Although this is less likely. What is more likely to be the reason behind this good performance is the language modeling/generating capability of GPT-2.

In the next step, we integrated visual information into LLM. Taking inspiration from \cite{clipcap}, we learn a projector network to align the visual features with GPT-2 representations (VisualGPT-2 in \autoref{tab:res_oe_consolidated}. We observed an improvement in the performance. However, we believe that this might not be a very efficient way to integrate visual information with LLMs. We expect the performance to boost considerably through better and more sophisticated joint modeling vision and language. Overall, we believe shallower networks might not have the capacity to do inference over longer, complex causal chains, and as such it might not be the best option to invest future efforts into; LLMs have the potential to excel at causal reasoning; this view is also supported in a concurrent survey \cite{liu2024large}.

\begin{wraptable}[9]{R}{0.3\columnwidth}
\vspace{-.5cm}
\resizebox{\columnwidth}{!}{%
\small
\centering
\setlength\tabcolsep{2pt}
\begin{tabular}{@{}lcccc@{}}
\toprule
\multirow{2}{*}{\textbf{\begin{tabular}[c]{@{}l@{}}Train\\ Data\end{tabular}}} & \multicolumn{3}{c}{\textbf{MCQA}} & \textbf{OEAG} \\ \cmidrule(lr){2-4} \cmidrule(l){5-5} 
       & \textbf{BlindQA} & \textbf{HGA}   & \textbf{MIST}  & \textbf{HGA}  \\ \cmidrule(r){1-1} \cmidrule(lr){2-2} \cmidrule(lr){3-3} \cmidrule(lr){4-4} \cmidrule(l){5-5}
NextQA & 32.73            & 41.23          & 55.44          & 1.44          \\
\textbf{+} Ours & \textbf{32.91}   & \textbf{41.44} & \textbf{55.96} & \textbf{1.48} \\ \bottomrule
\end{tabular}
}
\caption{\textbf{Our dataset improves performance on existing real-world dataset.} For OEAG, WUPS score is reported.}
\label{tab:res_xfer_mcqa}
\end{wraptable}
\subsection{Does our dataset help with real-world cases?}
To evaluate the direct impact of our dataset, we combine our training set with NextQA's training set and measure model performance on the NextQA test set. For comparison, we also measure model performance without incorporating our dataset.
Despite being a \textit{synthetic/cartoon} dataset, and \textit{much smaller in size}, our dataset boosted the performance on a real-world dataset on both MCQA and OEAG (\autoref{tab:res_xfer_mcqa}). We observed \textit{improvement in identifying the correct causes by breaking the reliance on shortcuts and focus on causal effects}; a more comprehensive analysis of the situation/interaction, rather than jumping to a conclusion. Qualitative results provided in \supplementary. What is more, improvements were not limited to `causal-why' questions but also extended across other question types. We hypothesize that this can attributed to the wider range of reasoning involved in our dataset. To the best of our knowledge, this is the first time a synthetic VideoQA dataset has shown immediate improvement on a real-world dataset. However, we noticed a slight drop in the performance on location-type questions. Thus, we believe that it might be better to: \textbf{1)} \textit{transfer reasoning skills acquired from challenging synthetic datasets like ours}; \textbf{2)} leverage our synthetic dataset to \textit{inform the model designing process}, as it better reflects the challenges VideoQA models may face in the real world, such as longer causal chains and frequent scene changes in visual streams. Nonetheless, we \textit{do not suggest naively deploying whole models/weights trained on our dataset to real-world scenarios or applications}.
\section{Conclusion}
\label{sec:conclusion}
We introduced CausalChaos!, a challenging dataset for causal action question-answering tasks based on the classic Tom and Jerry cartoon, richly annotated with critical thinking questions requiring extensive reasoning from Video QA models. Questions come with multi-level answers and explanations covering the entire video context. We also provide the novel CausalConfusion test set to challenge causal relationship modeling in Video QA models. Our experiments show that while existing models perform well on causal action QA tasks, there is significant room for improvement in causal relationship modeling and generating detailed open-ended answers. LLMs show promise, but integrating visual information with LLMs or joint modeling of vision and language is crucial. We hope our dataset fosters such developments and will release it and the codes to support future efforts. Lastly, we demonstrated improvements in real-world datasets. Our dataset, derived from cartoons, should inform model design, reflecting real-world challenges like longer causal chains and frequent scene changes. However, we do not suggest deploying models trained solely on our dataset in real-world scenarios.

\paragraph{Material Acknowledgement and Disclaimer.}
Tom and Jerry is a material of Turner Entertainment Company (Warner Bros. Entertainment Inc.). All rights reserved. We do not claim any ownership of or rights to the Tom and Jerry material. All other trademarks, service marks, trade names and any other material referenced in this document are the property of their respective owners.

\paragraph{Acknowledgements.}
This research/project is supported by the National Research Foundation, Singapore, under its NRF Fellowship (Award\# NRF-NRFF14-2022-0001).

{
    \small
    \bibliographystyle{ieeenat_fullname}
    \bibliography{main}
}

\end{document}